# Journal of Engineering Technology and Applied Physics

# Neural Network Facial Authentication for Public Electric Vehicle Charging Station

Muhamad Amin Husni Abdul Haris* and Sin Liang Lim
*Faculty of Engineering, Multimedia University, 63100 Cyberjaya, Malaysia.*
*Corresponding author*: aminhusni@gmail.com

*Abstract* – This study is to investigate and compare the facial recognition accuracy performance of Dlib ResNet against a K-Nearest Neighbour (KNN) classifier. Particularly when used against a dataset from an Asian ethnicity as Dlib ResNet was reported to have an accuracy deficiency when it comes to Asian faces. The comparisons are both implemented on the facial vectors extracted using the Histogram of Oriented Gradients (HOG) method and use the same dataset for a fair comparison. Authentication of a user by facial recognition in an electric vehicle (EV) charging station demonstrates a practical use case for such an authentication system.

## I. INTRODUCTION

A facial recognition system is a tool used for identifying human faces from digital images or video frames and find matches from databases of faces. Using machine learning and neural network, the further enhancement could be achieved in the application of facial recognition. More accurate and faster performance of the application was aimed for this project. This paper focuses on the performance comparison between KNN and Dlib ResNet. The facial feature extraction was done with deep metric learning by extracting a 128-d vector (128 real-valued numbers) from an image [1].

Dlib ResNet has an accuracy limitation with Asian faces [2, 3]. This paper will investigate the difference in performance and accuracy between Dlib ResNet and KNN in recognizing Asian faces in an attempt to achieve better facial recognition accuracy.

Based on an issue raised on Github for Dlib, there are accuracy problems when it comes to Asian faces. The accuracy dropped from the Labelled Face In The Wild (LFW) benchmark of 99.38 % to 98.18 % [4]. The Dlib model utilizes a ResNet network with 29 convolution layers which is based on the ResNet-34 network from K. He *et. al.* [5].

To demonstrate the application of facial recognition proposed in this paper, the system is implemented into financially authenticated users for a simulated electric vehicle charging station. The function of the facial recognition system for the EV charging station is to recognize the user, to authenticate the user and to retrieve data regarding the user's allowable power usage for the charging session.

The main difference between KNN and Dlib is that KNN is using a machine learning algorithm whereas Dlib ResNet is a multi-layered neural network. KNN has the advantage of running efficiently in a practical sense with only a simple computer where Dlib ResNet would require a specialized processor to which is designed to run neural network applications such as the Nvidia Tensor capable graphics card to run efficiently. However, machine learning algorithms such as the KNN has the disadvantage of being inefficient when it is running with a large dataset since the algorithm needs to constantly revisit the dataset itself against the live-input data. A neural network does not require it to access the training dataset thus, making it a faster and more efficient method when paired with the proper NN processor hardware.

## II. Methodology

### A. Benchmarking Accuracy Performance Between Dlib ResNet and KNN Classifier

In this paper, the HOG method was used to determine facial landmarks and extract the 128-d vector from the face image. The facial features were extracted using the HOG method, which is then applied on both KNN and Dlib ResNet facial recognition neural network (NN). The HOG is a feature descriptor used in computer vision and image processing for object detection. The technique counts the frequency of gradient orientation in localized regions of an image. When trained with enough dataset, the HOG neutral network can determine the facial landmarks accurately and consistently across multiple samples of faces. Accuracy rate result was done with a test benchmark code which compares the accuracy between the two facial recognition methods. Both of them uses the same algorithm to get the benchmarking result.

A set of known labelled faces were used to train the KNN network. Another set of separate datasets that are not used in training were used for testing the neural network to identify known faces. The dataset of training data example can be seen in Fig. 1.

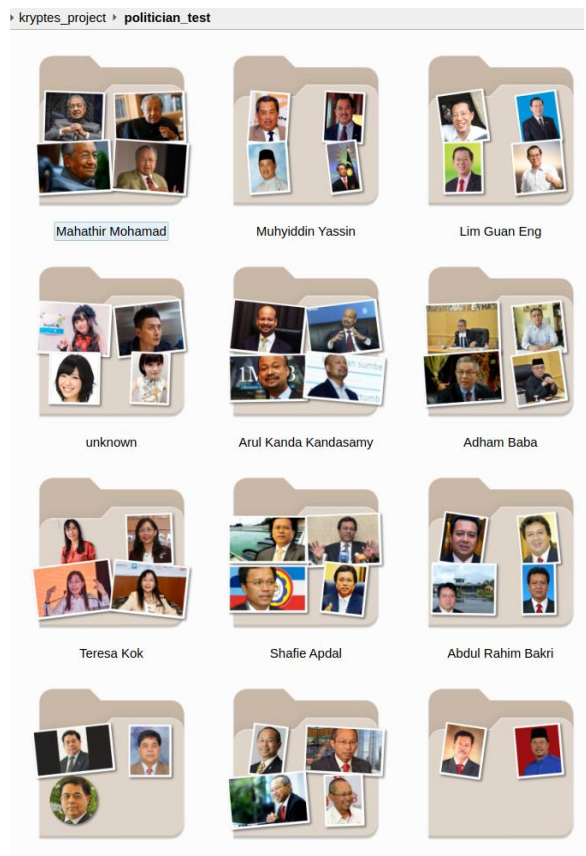

Fig. 1. Dataset of testing images for benchmarking.

Due to constraint of datasets, each training set has one to two images and there is a total of 15 datasets. The testing set has an average of 3 images. Each of them is sorted into folders with the labelled names. During the benchmarking process, the labelled names are noted by the algorithm and the facial recognition process was performed. The returned result is then compared with the label of the image. The benchmarking result for the accuracy comparisons between KNN and Dlib ResNet facial recognition was then obtained by calculating the amount of positive identification versus the testing data count. The overview of the steps taken and flow are summarized as follow: 1. Gather data (face data to be used for training and testing). 2. Filter and clean up the facial data. 3. Separate training and testing data. 4. Encode the facial data and train both with the same training data. 5. Run both facial recognition with the same testing data.

All of the training and test data are obtained from Sinar Project (Dataset License: CC BY-SA 4.0) application programming interface (API) and Microsoft Bing Image Search API (Result License Setting: Public Domain).

KNN and Dlib ResNet facial recognition is run against the same testing dataset with the same number and identical pictures to ensure a fair performance comparison. The returned result consists of:

Positive return – When the recognition matches the testing label
Negative return – When the recognition does not match with the testing label or unknown

The accuracy result is then calculated in percentages:

$$\frac{Positive\ Return}{Total\ Tested} * 100 = accuracy\ \% \qquad (1)$$

### B. Hardware Design

The energy meter current transformer clamp of the USB Modbus is attached to the live side of the 240V supply before the load. The status can then be read by the program by using the Modbus protocol. This is to simulate a vehicle charging from the station by consuming power.

A portable computer (eg. Raspberry Pi) is used to simulate an edge computing device. The facial feature is extracted from the camera on the edge device and the face vector data is transmitted to the server for facial recognition and matching with the users in the database. The hardware setup can be seen as shown in Fig. 2.

To simulate a charging vehicle and the function of electrical power quota mechanism, a dummy load was attached to the system as shown in Fig. 3. When the charging station is drawing power to the vehicle (dummy load), the power delivery is constantly monitored. When the power quota has met or exceeded for that particular user, the system will cut the power supply and end the charging session.

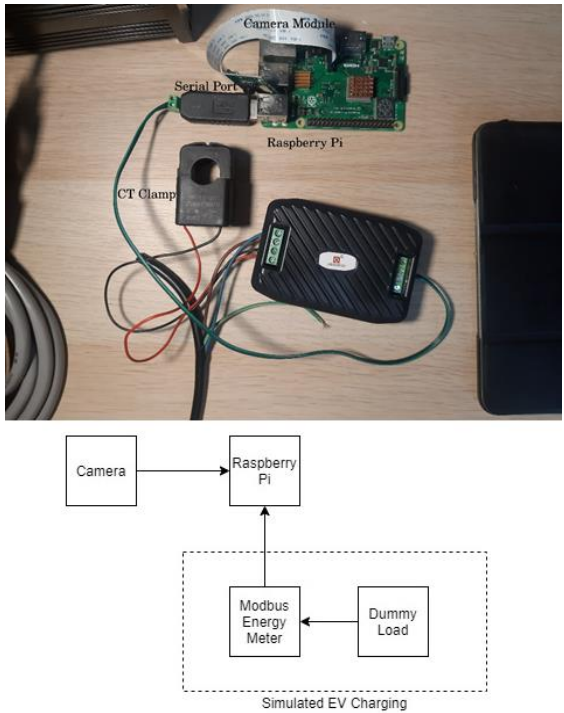

Fig. 2. Proof-of-concept prototype for the EV charging station.

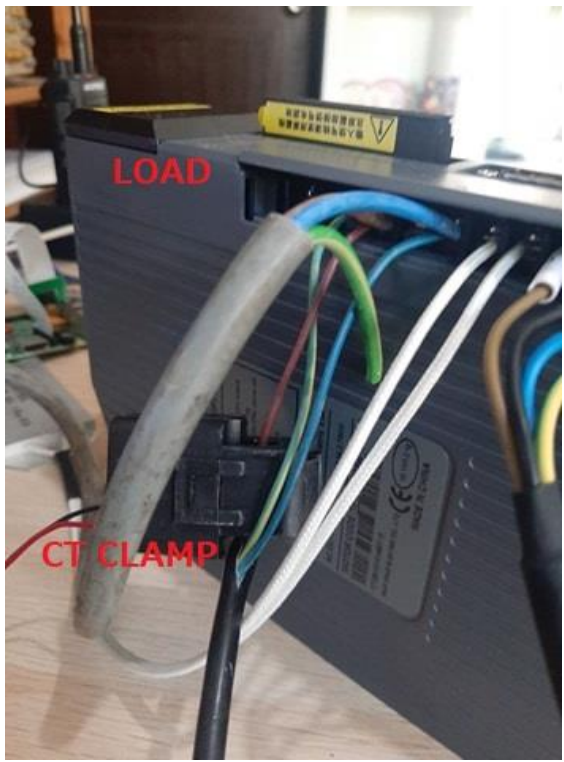

Fig. 3. Dummy load attached to the system to simulate a charging EV.

## III. RESULTS AND DISCUSSION

A total of 166 samples are used in the benchmarking program. Both facial recognition methods use the same dataset for a fair comparison of performance. HOG facial feature extraction method is used for both facial recognition methods.

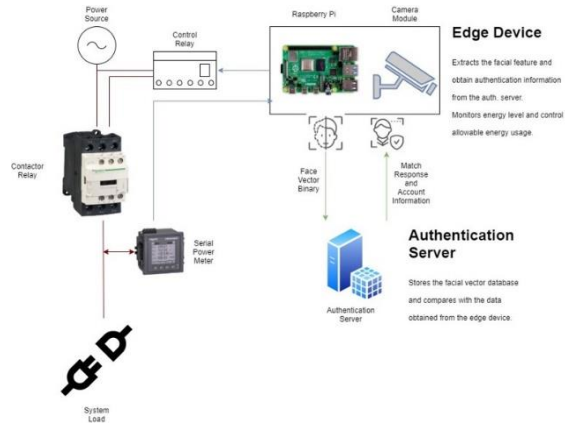

Fig. 4. Overall hardware design for edge computing facial authentication.

The final product is a prototype EV charging station that relies on facial recognition for authenticating and querying of data from the database to determine if the person is authorized and how much quota of electrical power is allocated to that person, as shown in Fig. 4. The overall network design is an edge-computing architecture. This allows the system to save bandwidth by transmitting only the extracted 128-d vector value to the authentication server instead of transmitting a full image or video. The 128-d facial feature is extracted by the edge device (EV station) on-site and the 128-d vector data is transmitted to the authentication server for facial recognition and authentication. The recognition and authentication of the face by using the vector are then done by the remote database server. The server will reply to the edge device (client) if the user is available and if there are charging wattage quota available for that particular user.

An example of 128-d facial feature for a person is shown in Fig. 5.

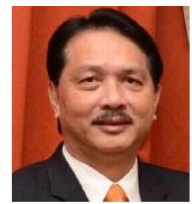

[-0.35,-0.15,...,0.35,0.42]

Fig. 5. Extraction of facial features.

The 128-d vector is then run against the facial recognition system to find a matching user in the database. When a matching user is found, the server will provide a return value which consists of the user information such as name, power consumption quota, account type, and so on. The program flow for EV charging station authentication is depicted in Fig. 6.

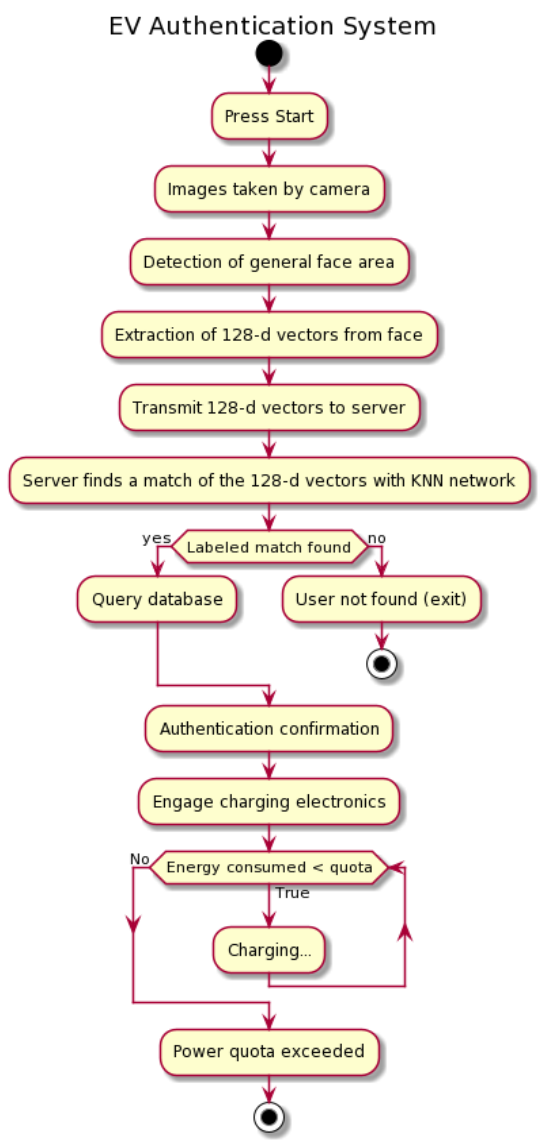

Fig. 6. Program flow for EV charging station authentication system.

In Fig. 7, the client-side will be handling the conversion of a picture into 128-d vector numbers. This will then be transmitted to the server in the form of a web request that expects a return of the user or a not found response.

*A. Accuracy Results*

KNN benchmark yields a total positive of 165 recognitions. The accuracy is calculated to be 98.78 %. Dlib ResNet benchmark yields a total positive of 144 recognitions. The accuracy is calculated to be 86.75 %.

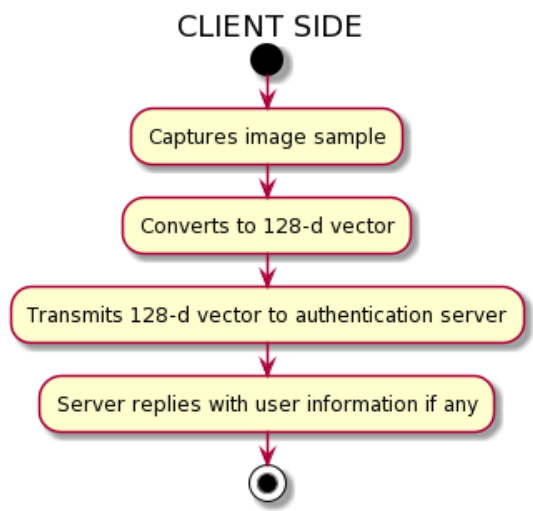

Fig. 7. Program flow on the client-side.

Table I. Accuracy results between the two methods.

| Method | Accuracy |
|---|---|
| KNN | 98.78 % |
| Dlib ResNet | 86.75 % |

*B. Data Discussion*

A set of parameters are tested for accuracy rate. This is the sensitivity setting for both the KNN and Dlib ResNet NN. The parameter with the highest accuracy rate is chosen for each facial recognition method.

KNN has yielded an accuracy of 98.8 % with 162 positive identification out of 166 test images.

Dlib ResNet has yielded an accuracy of 86.75 % with 144 positive identification out of 144 test images. As seen in Table I, it is found that facial recognition is more accurate when the KNN method is used compared to Dlib ResNet when it comes to geographically localized ethnicity in Malaysia which Dlib ResNet, is lacking the training data.

IV. CONCLUSION

It is found that the KNN network is more accurate when compared to the established ResNet-34 when recognizing Asian faces. A large and highly trained NN will not guarantee to work on a certain group of people, ethnicity in particular if it is not trained during the training process. This shows that the Dlib's ResNet will yield a low accuracy rate in recognizing Asian faces compared to recognizing western faces.

Such accuracy rate is not suitable to be used for financial authentication purposes as it will introduce false positives.

Dlib ResNet is still not suitable for use in Asian countries for the time being. More training is required with a new dataset of labelled faces containing a significant representation of Asian ethnicity. However, obtaining a good quality dataset for Asian faces would be a challenge [2].

## ACKNOWLEDGEMENT

The authors are grateful to C++ Dlib Project for advancing and putting academic research into a framework for practical use, and to all the developer contributors towards the Dlib project.